\documentclass[runningheads]{llncs}
\usepackage{graphicx}
\usepackage{amsmath,amssymb,amsfonts}
\usepackage{multirow}

\newcommand{\sysname}{LA-OOD}

\makeatletter
\newcommand{\multiline}[1]{%
  \begin{tabularx}{\dimexpr\linewidth-\ALG@thistlm}[t]{@{}X@{}}
    #1
  \end{tabularx}
}
\makeatother
\usepackage[normalem]{ulem}
\usepackage{algorithm}
\usepackage[noend]{algpseudocode}
\usepackage{hyperref}
\usepackage[dvipsnames]{xcolor}
\usepackage{wrapfig}

\begin{document}
\title{Layer Adaptive Deep Neural Networks for Out-of-distribution Detection}
%
%
\author{Haoliang Wang\inst{1} \and
Chen Zhao\inst{1} \and
Xujiang Zhao\inst{1} \and
Feng Chen\inst{1}}
\authorrunning{H. Wang et al.}
%
\institute{University of Texas at Dallas\\
\email{\{haoliang.wang, chen.zhao, xujiang.zhao, feng.chen\}@utdallas.edu}}
\maketitle              
\begin{abstract}
    During the forward pass of Deep Neural Networks (DNNs), inputs gradually transformed from low-level features to high-level conceptual labels. While features at different layers could summarize the important factors of the inputs at varying levels, modern out-of-distribution (OOD) detection methods mostly focus on utilizing their ending layer features. In this paper, we proposed a novel layer-adaptive OOD detection framework (\sysname{}) for DNNs that can fully utilize the intermediate layers’ outputs. Specifically, instead of training a unified OOD detector at a fixed ending layer, we train multiple One-Class SVM OOD detectors simultaneously at the intermediate layers to exploit the full-spectrum characteristics encoded at varying depths of DNNs. We develop a simple yet effective layer-adaptive policy to identify the best layer for detecting each potential OOD example. \sysname{} can be applied to any existing DNNs and does not require access to OOD samples during the training. Using three DNNs of varying depth and architectures, our experiments demonstrate that \sysname{} is robust against OODs of varying complexity and can outperform state-of-the-art competitors by a large margin on some real-world datasets.

\end{abstract}

\section{Introduction}
    Recently, deep neural networks (DNNs) have demonstrated remarkable performance in classification problems. However, DNNs are often designed for a static and closed world, assuming the same data distribution during training and test times. In an open-world environment, it is important to detect examples from novel class distributions in safety-critical applications (\textit{e.g.} detecting new categories of objects during autonomous driving and diagnoses of unknown diseases, such as COVID-19). It is hence necessary to develop DNNs that can identify OOD examples while at the same time classifying samples from known class distributions with high accuracy.  

A number of recent methods have been proposed to detect OOD examples based on DNNs. The majority of these methods detect OOD examples based on predictive uncertainty measures of a softmax classifier, such as entropy \cite{vyas2018out}, epistemic uncertainty \cite{DBLP:journals/corr/abs-1802-10501}, and others \cite{hendrycks2016baseline,liang2017enhancing,Zhao-ICDM-2019,zhao-KDD-2021}. A more recent work presents the Deep-MCDD \cite{lee2020multi}, that estimates a spherical decision boundary for each class based on support vector data description (SVDD), such boundaries will enclose the in-distribution (InD) samples and distinguish OODs based on their closest class-conditional distribution. Instead of using the last layer outputs, \cite{abdelzad2019detecting}  proposed to find the best intermediate layer based on a holdout validation OOD dataset.
However, all of the above methods detect the OOD examples at the same level of representation (\textit{i.e.} outputs at one single layer) and they hence fail to account for the different representation complexities of OOD examples. 
Particularly, our empirical study indicates that different OODs may be better detected at their appropriate levels of representations (see Section \ref{layerbehavior}).

\begin{figure*}[!t]
    \centering
    \includegraphics[width=0.9\linewidth]{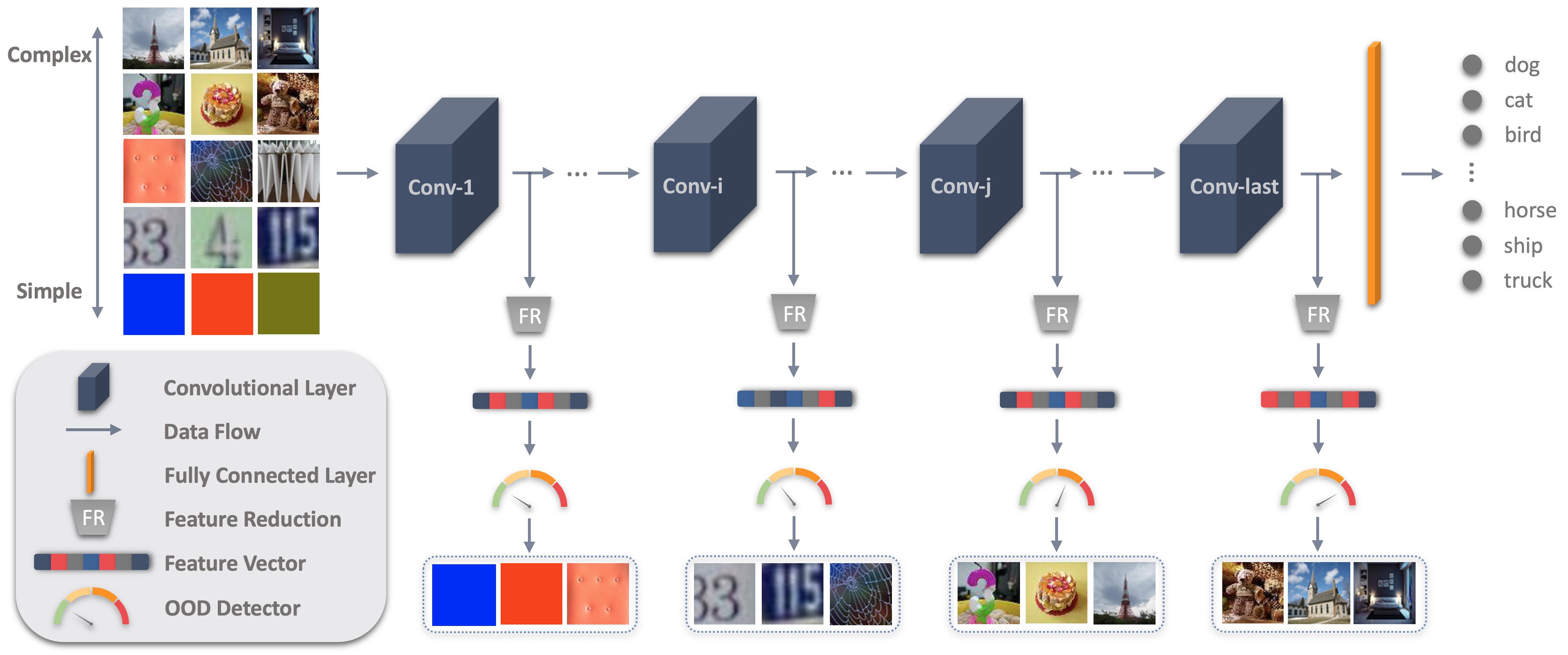}
    \vspace{-2mm}
    \caption{An overview of our proposed Layer Adaptive Deep Neural Networks for OOD Detection (\sysname{}).}
    \label{fig:overview}
    \vspace{-5mm}
\end{figure*}

This observation motivates us to propose a novel framework, namely Layer-Adaptive OOD detection (\sysname{}), a generic modification to off-the-shelf DNNs that introduces OOD detectors to intermediate layers. Specifically, we train separate One-Class SVM (OCSVM) OOD detectors using different layers' outputs and employ a simple yet effective layer-adaptive policy function to identify the best layer for detecting each potential OOD sample (see Figure \ref{fig:overview}). We tune the OOD detectors through self-adaptive data shifting \cite{wang2018hyperparameter} to improve its accuracy and robustness against unseen OODs, and fine tune the framework using alternating optimization, in which the DNN classification error and the OOD detectors' training errors are minimized jointly.

The main contributions are stated as follows:
\begin{itemize}
\vspace{-2mm}
    \item We propose a novel layer-adaptive OOD detection framework (\sysname{}) that is practical for any off-the-shelf DNNs. Multiple OOD detectors are attached to the intermediate layers of a DNN, through a simple yet effective layer-adaptive policy, our proposed framework is able to fully utilize the intrinsic characteristics of inputs encoded in the intermediate latent space, hence, detect OODs with varying complexity.
    \item We propose a joint objective that fine-tune the OOD detectors while maintaining DNN's classification accuracy. We also designed an OOD confusion metric and a Grad-CAM visualization tool to facilitate decision making and improve the model interpretability.
    \item Extensive experiments have been conducted to demonstrate the effectiveness of our proposed framework. On three DNNs with varying depth and architectures, using two InD datasets and five OOD datasets, \sysname{} outperform state-of-the-art baseline methods in most settings without any OOD training or validation samples, being a practical yet effective OOD detection framework for OODs of different complexity.
\end{itemize}

\section{Related Work}
    \textbf{Dynamic Neural Networks with Early-Exit.}
Adaptive early-exist is a rising research topic in deep learning. By attaching early exits to a DNN, such methods allow ``simple" samples to be output at early layers without ``overthinking" \cite{huang2017multi,kaya2019shallow}. For a given input, an early-exit could be determined by either a confidence metric \cite{leroux2017cascading} or a learned decision function \cite{bolukbasi2017adaptive}. However, these methods aim to improve DNN performance by focusing on InD sample evaluation without giving enough attention to OODs. 
\textit{In this paper, we adopt the idea of early exits for the out-of-distribution detection problem and propose a novel framework in which each OOD sample is detected at its best layer.}

\vspace{1mm}

\noindent \textbf{OOD Detection for Deep Neural Networks.}
In recent years, researchers have developed a number of OOD detection methods, where the majority of such techniques use the final outputs of a DNN to separate the OODs from the InD samples \cite{vyas2018out}. \cite{hendrycks2016baseline} proposes a baseline method that detects OODs based on the maximum softmax probabilities of a DNN's final outputs. ODIN \cite{liang2017enhancing} incorporates the temperature scaling and input perturbation into the maximum softmax probabilities to enhance the margin between InD and OOD samples. More recently, \cite{lee2020multi} extends Deep-SVDD to a multi-class setting and proposes the Deep-MCDD, It integrates multiple SVDDs into a single deep model where each SVDD is trained to surround one InD class sample. However, these works mainly focus on the high-level conceptual features outputted by the ending layers of DNNs while ignoring the low-level representations at the intermediate layers, hence, may "overthink" the problem and fail on OODs of relatively low complexity.
\textit{In contrast, \sysname{} not only considers the ending layers' outputs but also takes the intermediate layers into consideration to generate more accurate OOD predictions.}

Two existing methods \cite{abdelzad2019detecting,lee2018simple} utilize intermediate outputs of a DNN for OOD detection. \cite{lee2018simple} defines the confidence score of input as a weighted average of the Mahalanobis distance to the closest class-conditional distribution at each layer, such weighting function is trained using an additional validation set. \cite{abdelzad2019detecting} proposes the OODL which decides an optimal discernment layer based on a holdout OOD dataset. Both methods require the OOD samples during the training, such OOD samples not only are hard to obtain in real-world applications, but also make the trained models susceptible to unseen OODs. 
\textit{In this work, we tune the OOD detectors using pseudo OODs generated through self-adaptive data shifting \cite{wang2018hyperparameter} of the InD training samples, hence, does not require any OOD samples during the training.}

\section{Adaptive One-Class Deep Neural Network}

Since OOD samples are rarely available during the training, here we formulate the OOD detection as a one-class classification problem, in which OOD detectors only target to determine whether an input is in-distribution or not.

\subsection{Problem Formulation}
Let $\mathbf{x} \in \mathcal{X}$ be an input, $y \in \mathcal{Y}=\{1,\cdots,K\}$ being its label, given a deep neural network $\mathcal{M}$ with $L$ layers, it tries to classify each input to $K$ classes: $\hat{y}=\mathcal{M}(\mathbf{x})\in\mathcal{Y}$. With the intermediate outputs $\mathbf{x}^{(\ell)}$ at layer $\ell \in \{1, \cdots, L\}$, its OOD score $s^{(\ell)}=C_{\ell}(\mathbf{x}^{(\ell)})$ is computed by a layer-specific OOD detector $C_{\ell}$. Separate OOD detectors could be attached to different layers of $\mathcal{M}$, the final OOD score of $\mathbf{x}$ could be obtained by taking the maximum OOD scores outputted by all the OOD detectors: $s_{\text{final}} = \max \big[\{C_\ell(\mathbf{x}^{(\ell)})\}_{\ell=1}^L \big]$. Such OOD score then can be used to determine whether $\mathbf{x}$ is in-distribution or not based on a predefined threshold $\delta$.

\subsection{Framework Overview}

In the context of one-class classification, there are many possible selections for the OOD detector (KDE, GMM, $k$-NN, \textit{etc.}) In this paper, we use the One-Class Support Vector Machine (OCSVM) \cite{scholkopf2000support} which is one of the most commonly used one-class classifier in the literature. Note that, we could replace OCSVM with any other one-class classifiers as our framework design does not depend on a specific choice of one-class classifiers.

For the OCSVM, a feature mapping $\Phi:\mathcal{X}\subset\mathbb{R}^d\rightarrow \mathcal{F}\subset\mathbb{R}^h$ is defined, where $h>d$, it maps the input samples $\{\mathbf{x}_i\}_{i=1}^n\in\mathbb{R}^d$ into a high dimensional feature space $\mathcal{F}$. An OCSVM will try to find the best separating hyperplane that separates all the input samples from the origin such that the distance to the origin is maximized. Normally, the calculation of the feature mapping $\Phi$ is avoided by using the kernel trick $k(\mathbf{x}_i,\mathbf{x}_j)=(\Phi(\mathbf{x}_i)\cdot\Phi(\mathbf{x}_j))$. In this paper, we select the commonly used Gaussian Radial Base Function (RBF) kernel: $k\left(\mathbf{x}_i, \mathbf{x}_j\right)=\exp \left(-\gamma\left\|\mathbf{x}_i-\mathbf{x}_j\right\|^{2}\right)$, where $\gamma$ is the kernel width.

Using Lagrange multipliers, optimizing the OCSVM $C_\ell$ at layer $\ell$ is equivalent to solving the following dual Quadratic Programming (QP) problem:
\begin{align} \label{eq:SVMObjective}
    \min _{\mathbf{\boldsymbol{\alpha}^{(\ell)}}} \quad \frac{1}{2} \sum_{i,j} \boldsymbol{\alpha}_{i}^{(\ell)} \boldsymbol{\alpha}_{j}^{(\ell)} k\left(\mathbf{x}_{i}^{(\ell)}, \mathbf{x}_{j}^{(\ell)}\right) \quad
    \text {s.t. } 0 \leq \boldsymbol{\alpha}_{i}^{(\ell)} \leq \frac{1}{\nu n}, \text {and } \sum_{i} \boldsymbol{\alpha}_{i}^{(\ell)}=1 
\end{align}
where $\boldsymbol{\alpha}_i^{(\ell)}$ are the Lagrange multipliers, and $\nu\in(0,1]$ is the upper bound of the training error.

Given an input sample $\mathbf{x}$ and its layer $\ell$ outputs $\mathbf{x}^{(\ell)}$, its OOD score at layer $\ell$ is calculated using the decision function:
\begin{equation} \label{eq:4}
C_\ell(\mathbf{x})=- \sum_{i} \boldsymbol{\alpha}_{i}^{(\ell)} k\left(\mathbf{x}_{i}^{(\ell)}, \mathbf{x}^{(\ell)}\right)+\rho^{(\ell)}
\end{equation}
where the offsets $\rho^{(\ell)}$ can be recovered by $\rho^{(\ell)}=\sum_{j} \boldsymbol{\alpha}_{j}^{(\ell)} k\left(\mathbf{x}_{j}^{(\ell)}, \mathbf{x}_{i}^{(\ell)}\right)$. Positive scores represent OODs, and negative scores represent InDs (assuming the default zero threshold is used, \textit{i.e.}, $\delta=0$).

\subsection{Framework Training} \label{section:framework_training}

Given a pre-trained DNN model $\mathcal{M}_{\boldsymbol{\theta}}$ parameterized by $\boldsymbol{\theta}$, using the OCSVMs as OOD detectors, we propose a joint objective for training both the backbone model and the OOD detectors:
\begin{align}
\label{eq:training-model}
    \underset{\boldsymbol{\theta}}{\min} \underset{{\boldsymbol{\alpha}^{(\ell)}}_{\ell = 1}^L}{\min} \quad &L(\boldsymbol{\theta}) + \frac{\lambda}{2} \cdot \sum_{\ell=1}^L  \sum_{i,j} \boldsymbol{\alpha}_{i}^{(\ell)} \boldsymbol{\alpha}_{j}^{(\ell)} k\left(\mathbf{x}_{i}^{(\ell)}, \mathbf{x}_{j}^{(\ell)}\right) \\
    \text {subject to } \quad &0 \leq \boldsymbol{\alpha}_{i}^{(\ell)} \leq \frac{1}{\nu n}, \text {and } \sum_{i} \boldsymbol{\alpha}_{i}^{(\ell)}=1 \nonumber
\end{align}
Here the first term $L(\boldsymbol{\theta})$ denotes the loss function of the backbone network, and the second term is the summation of losses for all the OOD detectors multiplied by a regularization parameter $\lambda>0$. We aim to fine-tune the layer-dependent feature representations and the parameters of layer-dependent OCSVM jointly so that the training errors of the OOD detectors are minimized while maintaining DNN's classification accuracy.

To solve Eq.(\ref{eq:training-model}), an alternating optimization technique is applied in which the $\boldsymbol{\theta}$ and $\{\boldsymbol{\alpha}^{(\ell)}\}_{\ell=1}^L$ will be updated alternatively:

\begin{itemize}
    \item Step I: Fix $\{\boldsymbol{\alpha}^{(\ell)}\}_{\ell=1}^L$ and re-estimate the model parameters $\boldsymbol{\theta}$ using a Eq. \ref{eq:backbone_loss}.
    \item Step II: Fix $\boldsymbol{\theta}$ and generate the updated intermediate outputs to re-estimate $\{\boldsymbol{\alpha}^{(\ell)}\}_{\ell=1}^L$ using Eq. \ref{eq:SVMObjective}.
\end{itemize}

In step I, we fix the estimated dual coefficients $\{\boldsymbol{\alpha}^{(\ell)}\}_{\ell=1}^L$ for all OCSVMs, then re-estimate the backbone model parameter $\boldsymbol{\theta}$:
\begin{eqnarray} \label{eq:backbone_loss}
\min_{\boldsymbol{\theta}} \quad L(\boldsymbol{\theta}) + \frac{\lambda}{2L} \sum_{\ell=1}^L \sum_{i, j} \boldsymbol{\alpha}_i^{(\ell)} \boldsymbol{\alpha}_j^{(\ell)} k({\bf x}_i^{(\ell)}(\boldsymbol{\theta}), {\bf x}_j^{(\ell)}(\boldsymbol{\theta}))
\end{eqnarray}
In step II, we fix the backbone model to update the intermediate outputs for the training samples, then based on the newly generated outputs, we re-train all the OOD detectors using Eq. \ref{eq:SVMObjective}.

Two important hyper-parameters for OCSVM training are the Gaussian kernel width $\gamma$ and the training error upper bound $\nu$. $\gamma$ controls the smoothness of the decision boundary. The smaller the $\gamma$, the smoother the decision boundary will be. $\nu$ controls the error ratio, which is often tuned to reject the noisy samples in the training set and it also determines a lower bound on the fraction of support vectors. These two hyper-parameters are critical for OCSVM to achieve good performance. In general, these hyper-parameters are tuned using a held-out validation set that includes both InD and OOD samples. 
In this work, we adopt the self-adaptive data shifting \cite{wang2018hyperparameter} to generate pseudo-OODs for hyper-parameter tuning. Such pseudo-OODs are created purely using InD samples through edge pattern detection \cite{li2010selecting}.
We summarized our \sysname{} training procedure in Algorithm \ref{alg:LA-OOD}.

\subsection{Layer-Adaptive Policy Design}
Having $L$ OCSVM OOD detectors $\{C_\ell\}_{\ell=1}^L$ that each outputs an OOD score $s^{(\ell)}_i$ for input $\mathbf{x}_i$, we either need to define a threshold for each of these OOD detectors or design a decision policy that consolidates all the OOD scores into a final prediction. Empirically, we found that a layer-adaptive policy performs better than some fixed thresholds as it is very common that the predictions of OOD detectors diverge from each other (see Section \ref{section:confusion}). Here we choose a simple yet effective layer-adaptive policy that propagates the most confident opinion among all OOD detectors as the final prediction, specifically, the policy is design as $s_{i,\text{final}} = \max \big[\{C_\ell(\mathbf{x}^{(\ell)}_i)\}_{\ell=1}^L \big]$. One challenge to such policy design is that OCSVMs trained on different features generally will have a different scale of scores, this effect could be alleviated by normalizing the training features for each OCSVM, here we simply use the standardization: $\mathbf{x}^{'}=(\mathbf{x}-\mathbf{\Bar{x}})/\sigma$, with $\Bar{\mathbf{x}}$ being the sample mean and $\sigma$ being its standard deviation.

\begin{algorithm}[t]
    \caption{LA-OOD Training Procedure}\label{alg:LA-OOD}
    \textbf{Input:}  Pre-trained DNN model $\mathcal{M}_{\boldsymbol{\theta}}$, InD sample set $\mathcal{X}$ \\
    \textbf{Output:} Jointly trained $\mathcal{M}_{\boldsymbol{\theta}}$ and OOD detectors $\{C_\ell\}_{\ell=1}^L$
\begin{algorithmic}[1]
    \State Generate the intermediate outputs $\{\mathcal{X}^{(\ell)}\}_{\ell=1}^L$ 
    \State Generate pseudo-outliers
    \Statex $\{\mathcal{X}^{(\ell)}_{\text{pseudo}}\}_{\ell=1}^L=\text{selfAdaptiveDataShifting}(\{\mathcal{X}^{(\ell)}\}_{\ell=1}^L)$
    \State Hyper-parameter tuning for $\{C_\ell\}_{\ell=1}^L$ using $\{\mathcal{X}^{(\ell)}\}_{\ell=1}^L$ and $\{\mathcal{X}^{(\ell)}_{\text{pseudo}}\}_{\ell=1}^L$
    \While{not done}
        \State Fix the $\{\boldsymbol{\alpha}^{(\ell)}\}_{\ell=1}^L$ and re-estimate $\boldsymbol{\theta}$ (Eq. \ref{eq:backbone_loss})
        \State Update the intermediate outputs $\{\mathcal{X}^{*(\ell)}\}_1^{L}$
        \State Re-train $\{C_\ell\}_{\ell=1}^L$ using the updated intermediate outputs 
        \Statex $\{\mathcal{X}^{*(\ell)}\}_1^{L}$ (Eq. \ref{eq:SVMObjective})
    \EndWhile
    \State \textbf{return} trained $\mathcal{M}_{\boldsymbol{\theta}}$ and $\{C_\ell\}_{\ell=1}^L$
\end{algorithmic}
\end{algorithm}

\section{Experimental Results}
    
\textbf{Empirical Settings\footnote{The source code and datasets are available at: \href{https://github.com/haoliangwang86/LA-OOD}{https://github.com/haoliangwang86/LA-OOD}}.} 
(1) \textbf{Datasets.}
Two InD datasets (CIFAR10 and CIFAR100) and five OOD datasets (LSUN, Tiny ImageNet, SVHN, DTD\cite{cimpoi2014describing}, and Pure Color) are considered in the experiments. The ``Pure Color" dataset is a synthetic dataset that contains 10,000 randomly generated pure-color images. 
For each InD-OOD combination, we construct a training set using all the training images in the InD dataset and form a balanced test set using all the test images in both InD and OOD datasets, when the sizes of their test set mismatch, we randomly selected the same number of images from the larger dataset to match the test sample size of the smaller one. All images are down-sampled to $32 \times 32$ resolution using Lanczos interpolation.
(2) \textbf{Backbone Models.}
We evaluate our method using three popular CNNs in computer vision and machine learning studies. Particularly, we select the VGG-16, ResNet-34, and DenseNet-100 to demonstrate the effectiveness of our framework for DNN models of varying depth and architectures.
(3) \textbf{Feature Reduction.}
A feature reduction operation is applied to the intermediate outputs to maintain the scalability\cite{abdelzad2019detecting}. Among the pooling methods we have tested: max/average pooling with various sizes, global max/average pooling, the global average pooling performs the best. The pooled features are then standardized using the training set mean and deviation.
(4) \textbf{Hyper-parameters Tuning.}
We fix $\nu$ to be $0.001$ so that only a small number of InD samples will be considered as noise, the $\gamma$ is tuned using pseudo-OODs generated by self-adaptive data shifting \cite{wang2018hyperparameter} of only the InD training samples. We search $\gamma$ in $[0.001, 0.0025, 0.005, 0.01, 0.025, 0.05, 0.1, 0.25, 0.5, 1.0]$, for different InD-backbone pairs, we will shrink the value range to accommodate the differences in feature complexity and to reduce training time.
(4) \textbf{Baseline Methods and Evaluation Metrics.}
We compare our method with four state-of-the-art OOD detection baselines: MSP\cite{hendrycks2016baseline}, ODIN\cite{liang2017enhancing} (both temperature scaling and input preprocessing are used to achieve optimum performance), OODL\cite{abdelzad2019detecting} (we use the iSUN \cite{xu2015turkergaze} as an additional OOD dataset to find its optimal discernment layer), and Deep-MCDD\cite{lee2020multi}. Three commonly adopted OOD detection metrics are used: AUROC, AUPR, and FPR at 95\% TPR.


\subsection{Performance Evaluation} \label{section:43}

The experimental results are reported in Table \ref{table:performance_evaluation}, the mean values of the each evaluation metric are also reported to demonstrate the overall performance on OOD datasets with varying complexities. It is worth noting that previous works often choose to use linear interpolation for the down-sampling operation\cite{abdelzad2019detecting,lee2020multi,lee2018simple,liang2017enhancing}, however, we found that \textit{using linear interpolation will create severe aliasing artifacts which make such OOD samples easily detectable}, therefore, to generate more genuine OOD samples, we down-sampled the OOD images using the Lanczos interpolation which is much more sophisticated than the linear interpolation.

From Table \ref{table:performance_evaluation}, it could be seen that OODs that of higher complexity will be harder to detect, such as the LSUN and Tiny ImageNet images that could contain complex backgrounds or multiple objects in a single image. OODs of lower complexity are easier to detect, such as the SVHN that contains cropped street view house numbers or DTD that contains images of different textures. The synthetic Pure Color dataset is of the lowest complexity as it contains limited information. Such dataset complexity could be easily verified using entropy or energy metrics.

\begin{table*}[!t]
\normalsize
\centering
\caption{Performance evaluation. Metrics with "$\uparrow$" indicate the bigger the better and "$\downarrow$" indicate the smaller the better. Best performance are labeled in \textbf{bold}.}
\setlength\tabcolsep{3.5pt}
\resizebox{\textwidth}{!}{
\begin{tabular}{ccc|c|c}
\hline
\multirow{2}{*}{InD   / Model}                                                  & \multirow{2}{*}{OOD}              & AUROC $\uparrow$                        & AUPR $\uparrow$ & FPR at 95\% TPR  $\downarrow$\\ \cline{3-5} 
                                                                                &      &\multicolumn{3}{c}{MSP / ODIN /   Deep-MCDD / OODL / \sysname{} (Ours)}                                                                                                          \\ \hline
\multirow{6}{*}{\begin{tabular}[c]{@{}c@{}}Cifar10\\      VGG-16\end{tabular}}        & LSUN                 & \multicolumn{1}{c|}{86.25 / 86.75 / 85.19 / \textbf{88.03} / 87.26} & \multicolumn{1}{c|}{85.26 / 87.06 / 84.76 / \textbf{88.01} / 84.42} & 69.27 / 67.72 / 59.09 / 62.38 / \textbf{54.88} \\
                                                                                      & Tiny                 & \multicolumn{1}{c|}{85.66 / 86.35 / 83.95 / 87.10 / \textbf{88.39}} & \multicolumn{1}{c|}{84.23 / 86.22 / 83.49 / \textbf{86.98} / 86.02} & 67.36 / 64.30 / 61.56 / 64.08 / \textbf{44.00} \\
                                                                                      & SVHN                 & \multicolumn{1}{c|}{91.12 / 91.47 / 89.81 / 91.68 / \textbf{97.27}} & \multicolumn{1}{c|}{87.06 / 89.29 / 93.99 / 88.46 / \textbf{97.15}} & 21.78 / 25.45 / 64.02 / 23.52 / \textbf{14.25} \\
                                                                                      & DTD                  & \multicolumn{1}{c|}{87.73 / 90.26 / 88.33 / 92.16 / \textbf{97.35}} & \multicolumn{1}{c|}{87.05 / 89.58 / 80.60 / 90.82 / \textbf{97.45}} & 66.24 / 46.33 / 53.56 / 25.04 / \textbf{14.06} \\
                                                                                      & Pure Color           & \multicolumn{1}{c|}{98.57 / 99.77 / 98.42 / 99.41 / \textbf{99.93}} & \multicolumn{1}{c|}{98.18 / 99.75 / 98.30 / 98.94 / \textbf{99.84}} & 04.66 / 01.24 / 05.68 / 02.08 / \textbf{00.21} \\ \cline{2-5} 
                                                                                      & \textbf{Mean}                 & \multicolumn{1}{c|}{89.87 / 90.92 / 89.14 / 91.68 / \textbf{94.04}} & \multicolumn{1}{c|}{88.36 / 90.38 / 88.23 / 90.64 / \textbf{92.98}} & 45.86 / 41.01 / 48.78 / 35.42 / \textbf{25.48} \\ \hline
\multirow{6}{*}{\begin{tabular}[c]{@{}c@{}}Cifar100\\      VGG-16\end{tabular}}       & LSUN                 & \multicolumn{1}{c|}{73.00 / 73.58 / 72.83 / \textbf{75.10} / 72.48} & \multicolumn{1}{c|}{68.49 / 69.78 / \textbf{69.92} / 69.68 / 65.28} & 75.43 / \textbf{74.92} / 85.12 / 74.99 / 80.24 \\
                                                                                      & Tiny                 & \multicolumn{1}{c|}{77.10 / 77.83 / 76.37 / 79.84 / \textbf{80.57}} & \multicolumn{1}{c|}{72.64 / 74.82 / 73.27 / \textbf{75.20} / 75.19} & 63.53 / 68.89 / 80.50 / 60.68 / \textbf{56.22} \\
                                                                                      & SVHN                 & \multicolumn{1}{c|}{75.43 / 78.18 / 74.98 / 78.43 / \textbf{87.07}} & \multicolumn{1}{c|}{71.53 / 76.20 / 86.52 / 72.63 / \textbf{85.82}} & 66.26 / 70.29 / 82.31 / 62.78 / \textbf{48.94} \\
                                                                                      & DTD                  & \multicolumn{1}{c|}{75.75 / 76.81 / 73.80 / 77.76 / \textbf{93.28}} & \multicolumn{1}{c|}{70.20 / 72.94 / 58.84 / 70.63 / \textbf{93.33}} & 62.13 / 64.66 / 82.20 / 57.82 / \textbf{33.20} \\
                                                                                      & Pure Color           & \multicolumn{1}{c|}{62.66 / 51.22 / 78.28 / 58.10 / \textbf{96.71}} & \multicolumn{1}{c|}{54.24 / 49.93 / 73.44 / 49.13 / \textbf{95.24}} & 72.32 / 95.31 / 81.83 / 64.85 / \textbf{30.08} \\ \cline{2-5} 
                                                                                      & \textbf{Mean}                 & \multicolumn{1}{c|}{72.79 / 71.52 / 75.25 / 73.85 / \textbf{86.02}} & \multicolumn{1}{c|}{67.42 / 68.73 / 72.40 / 67.45 / \textbf{82.97}} & 67.93 / 74.81 / 82.39 / 64.22 / \textbf{49.74} \\ \hline
\multirow{6}{*}{\begin{tabular}[c]{@{}c@{}}Cifar10\\      ResNet-34\end{tabular}}     & LSUN                 & \multicolumn{1}{c|}{90.16 / 90.26 / 88.02 / \textbf{91.97} / 89.06} & \multicolumn{1}{c|}{87.62 / 90.19 / 86.74 / \textbf{90.56} / 84.48} & 33.24 / 50.28 / 55.75 / \textbf{31.19} / 37.35 \\
                                                                                      & Tiny                 & \multicolumn{1}{c|}{86.53 / 85.46 / 83.34 / 88.81 / \textbf{89.29}} & \multicolumn{1}{c|}{84.79 / 86.46 / 83.25 / \textbf{87.66} / 86.47} & 58.26 / 74.41 / 61.28 / 46.15 / \textbf{36.90} \\
                                                                                      & SVHN                 & \multicolumn{1}{c|}{84.33 / 81.22 / 88.08 / 87.74 / \textbf{97.77}} & \multicolumn{1}{c|}{81.88 / 81.89 / 93.97 / 85.13 / \textbf{97.67}} & 66.58 / 81.16 / 57.06 / 42.84 / \textbf{12.17} \\
                                                                                      & DTD                  & \multicolumn{1}{c|}{87.64 / 83.96 / 84.56 / 92.10 / \textbf{97.91}} & \multicolumn{1}{c|}{85.24 / 84.39 / 75.07 / 91.10 / \textbf{98.06}} & 51.61 / 78.01 / 62.13 / 30.82 / \textbf{11.84} \\
                                                                                      & Pure Color           & \multicolumn{1}{c|}{94.59 / 96.84 / 96.11 / 95.52 / \textbf{99.99}} & \multicolumn{1}{c|}{93.48 / 96.93 / 93.81 / 94.35 / \textbf{99.99}} & 17.84 / 15.54 / 36.80 / 19.50 / \textbf{00.04} \\ \cline{2-5} 
                                                                                      & \textbf{Mean}                 & \multicolumn{1}{c|}{88.65 / 87.55 / 88.02 / 91.23 / \textbf{94.80}} & \multicolumn{1}{c|}{86.60 / 87.97 / 86.57 / 89.76 / \textbf{93.33}} & 45.51 / 59.88 / 54.60 / 34.10 / \textbf{19.66} \\ \hline
\multirow{6}{*}{\begin{tabular}[c]{@{}c@{}}Cifar100\\      ResNet-34\end{tabular}}    & LSUN                 & \multicolumn{1}{c|}{75.63 / \textbf{77.52} / 74.65 / 51.91 / 65.25} & \multicolumn{1}{c|}{70.76 / \textbf{72.81} / 70.14 / 51.92 / 59.65} & \textbf{62.63} / 63.51 / 84.34 / 94.84 / 78.61 \\
                                                                                      & Tiny                 & \multicolumn{1}{c|}{78.70 / \textbf{81.28} / 78.29 / 67.05 / 75.82} & \multicolumn{1}{c|}{74.47 / 77.39 / \textbf{78.26} / 66.91 / 73.74} & 57.97 / \textbf{57.47} / 78.84 / 90.27 / 68.91 \\
                                                                                      & SVHN                 & \multicolumn{1}{c|}{78.76 / 84.16 / 78.62 / 79.00 / \textbf{84.61}} & \multicolumn{1}{c|}{73.71 / 78.74 / \textbf{88.50} / 69.18 / 76.09} & 55.29 / 46.58 / 77.50 / 45.81 / \textbf{36.85} \\
                                                                                      & DTD                  & \multicolumn{1}{c|}{75.32 / 78.94 / 77.11 / 86.25 / \textbf{91.39}} & \multicolumn{1}{c|}{70.07 / 74.52 / 84.85 / 83.45 / \textbf{91.97}} & 62.59 / 60.60 / 81.49 / \textbf{40.94} / 41.19 \\
                                                                                      & Pure Color           & \multicolumn{1}{c|}{55.23 / 62.25 / 63.47 / 96.46 / \textbf{99.80}} & \multicolumn{1}{c|}{48.09 / 52.11 / 53.16 / 91.14 / \textbf{99.78}} & 67.52 / 59.04 / 99.32 / 04.98 / \textbf{01.04} \\ \cline{2-5} 
                                                                                      & \textbf{Mean}                 & \multicolumn{1}{c|}{72.73 / 76.83 / 74.43 / 76.13 / \textbf{83.37}} & \multicolumn{1}{c|}{67.42 / 71.11 / 74.98 / 72.52 / \textbf{80.25}} & 61.20 / 57.44 / 84.30 / 55.37 / \textbf{45.32} \\ \hline
\multirow{6}{*}{\begin{tabular}[c]{@{}c@{}}Cifar10\\      DenseNet-100\end{tabular}}  & LSUN                 & \multicolumn{1}{c|}{92.07 / \textbf{94.01} / 87.19 / 88.47 / 84.38} & \multicolumn{1}{c|}{89.47 / \textbf{93.12} / 86.23 / 84.87 / 80.95} & 26.40 / \textbf{23.71} / 55.00 / 40.69 / 51.55 \\
                                                                                      & Tiny                 & \multicolumn{1}{c|}{89.96 / \textbf{91.95} / 85.22 / 84.62 / 88.75} & \multicolumn{1}{c|}{87.69 / \textbf{91.32} / 84.44 / 80.90 / 87.80} & 35.09 / \textbf{34.04} / 58.14 / 57.25 / 43.73 \\
                                                                                      & SVHN                 & \multicolumn{1}{c|}{89.00 / 89.54 / 89.48 / 97.19 / \textbf{97.79}} & \multicolumn{1}{c|}{85.73 / 88.11 / 94.46 / \textbf{97.54} / 97.51} & 36.33 / 43.54 / 51.29 / 16.07 / \textbf{09.41} \\
                                                                                      & DTD                  & \multicolumn{1}{c|}{88.65 / 85.42 / 86.93 / 95.10 / \textbf{97.61}} & \multicolumn{1}{c|}{86.06 / 84.75 / 77.33 / 96.14 / \textbf{97.58}} & 39.61 / 60.98 / 59.57 / 33.07 / \textbf{12.00} \\
                                                                                      & Pure Color           & \multicolumn{1}{c|}{91.83 / 96.78 / 96.21 / 79.15 / \textbf{99.97}} & \multicolumn{1}{c|}{87.80 / 95.01 / 95.08 / 69.92 / \textbf{99.97}} & 16.06 / 09.31 / 23.84 / 40.08 / \textbf{00.17} \\ \cline{2-5} 
                                                                                      & \textbf{Mean}                 & \multicolumn{1}{c|}{90.30 / 91.54 / 89.01 / 88.91 / \textbf{93.70}} & \multicolumn{1}{c|}{87.35 / 90.46 / 87.51 / 85.87 / \textbf{92.76}} & 30.70 / 34.32 / 49.57 / 37.43 / \textbf{23.37} \\ \hline
\multirow{6}{*}{\begin{tabular}[c]{@{}c@{}}Cifar100\\      DenseNet-100\end{tabular}} & LSUN                 & \multicolumn{1}{c|}{76.38 / \textbf{77.41} / 75.17 / 59.11 / 69.69} & \multicolumn{1}{c|}{72.14 / \textbf{73.19} / 71.18 / 57.10 / 64.28} & \textbf{62.62} / 65.02 / 82.93 / 91.64 / 72.59 \\
                                                                                      & Tiny                 & \multicolumn{1}{c|}{79.73 / \textbf{84.27} / 78.25 / 61.84 / 81.29} & \multicolumn{1}{c|}{76.10 / \textbf{81.66} / 75.11 / 59.22 / 78.81} & 55.24 / \textbf{50.97} / 77.48 / 81.85 / 62.76 \\
                                                                                      & SVHN                 & \multicolumn{1}{c|}{80.08 / 81.30 / 74.99 / 71.73 / \textbf{86.99}} & \multicolumn{1}{c|}{75.29 / 74.89 / \textbf{86.25} / 65.36 / 78.23} & 51.73 / 49.32 / 82.48 / 66.07 / \textbf{32.89} \\
                                                                                      & DTD                  & \multicolumn{1}{c|}{73.18 / 70.29 / 79.34 / 84.69 / \textbf{93.79}} & \multicolumn{1}{c|}{69.03 / 67.93 / 66.09 / 84.72 / \textbf{93.95}} & 73.09 / 91.60 / 75.11 / 56.15 / \textbf{30.67} \\
                                                                                      & Pure Color           & \multicolumn{1}{c|}{79.60 / 80.86 / 91.14 / 85.39 / \textbf{99.47}} & \multicolumn{1}{c|}{73.54 / 77.68 / 89.64 / 79.53 / \textbf{99.41}} & 44.87 / 61.26 / 49.77 / 34.72 / \textbf{02.84} \\ \cline{2-5} 
                                                                                      & \textbf{Mean}                 & \multicolumn{1}{c|}{77.79 / 78.83 / 79.78 / 72.55 / \textbf{86.25}} & \multicolumn{1}{c|}{73.22 / 75.07 / 77.65 / 69.19 / \textbf{82.94}} & 57.51 / 63.63 / 73.55 / 66.09 / \textbf{40.35} \\ \hline
\end{tabular}}
\label{table:performance_evaluation}
\vspace{-7mm}
\end{table*}

The OOD detection methods that utilize the ending layers' features (MSP, ODIN, and Deep-MCDD) generally perform well on detecting OODs with higher complexity, such as the LSUN and the Tiny ImageNet datasets, however, they tend to give poor decisions for OODs of lower complexity such as the SVHN, DTD, and the Pure Color datasets. The OODL baseline method could utilize the intermediate features, from the performance evaluation, we could see that OODL exhibit the same performance pattern as MSP, ODIN, and MCDD, however, it is due to that LSUN and Tiny ImageNet have similar complexity as the iSUN dataset, which is used to determine the optimal discernment layers for OODL, when the test OODs are of different complexity compare to iSUN, its performance could degrade significantly.

Through multiple intermediate OOD detectors and the layer-adaptive policy, \sysname{} can exploit the full-spectrum characteristics encoded in different intermediate layers. Specifically, by taking the early layers' outputs into consideration, \sysname{} outperforms the other four baseline methods by a large margin on OOD datasets of lower complexity (SVHN, DTD, and Pure Color). More importantly, \sysname{} achieves the best average AUROC/AUPR/FPR at 95\% TPR for all InD-Backbone settings, which indicates our proposed method is robust against OODs of different complexity. Overall, \sysname{} achieves an 8.21\% improvement margin on AUROC, 7.8\% improvement margin on AUPR, and 29.98\% improvement margin on FPR at 95\% TPR compare to the second-best baseline method.

\subsection{Understanding the Behaviors of Different Layers} \label{layerbehavior}

\begin{wrapfigure}{r}{0.49\linewidth}
    \vspace{-8mm}
    \includegraphics[width=\linewidth]{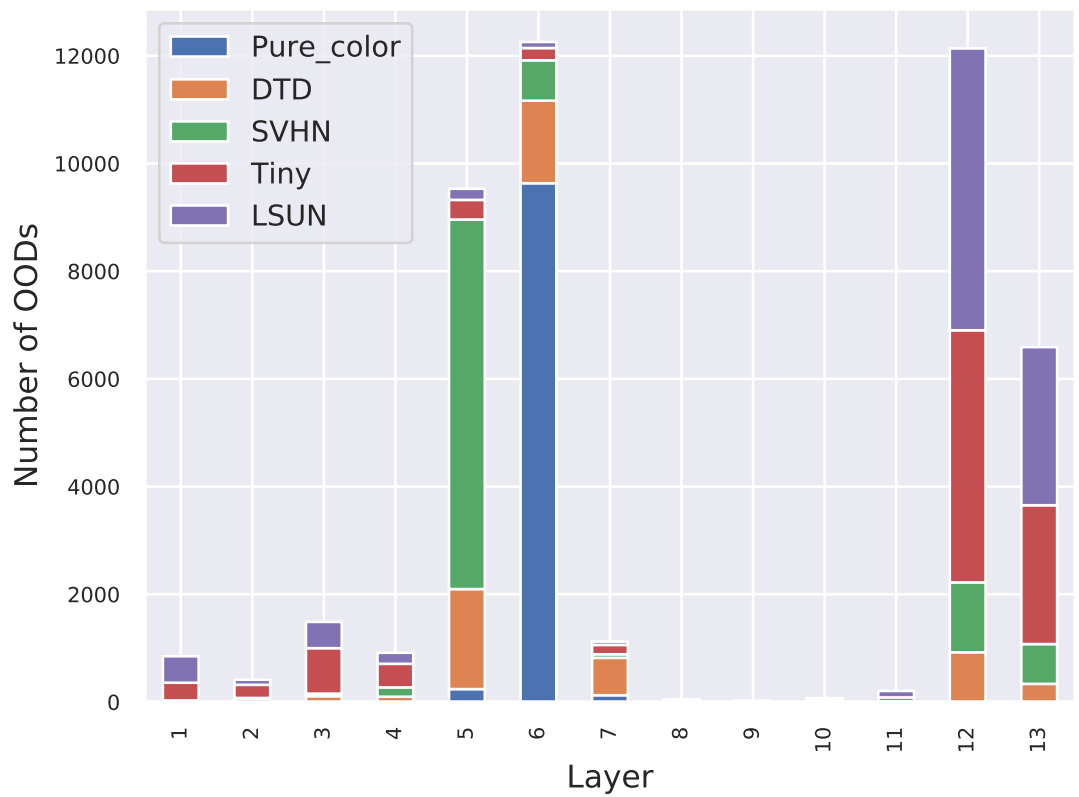}
    \vspace{-6mm}
    \caption{Number of OODs detected by OOD detectors at different layers using VGG-16 and CIFAR10 InD.}
    \label{fig:number_of_detected_ood}
    \vspace{-4mm}
\end{wrapfigure}
 
As the layer of a DNN goes deeper, more complex features could be learned\cite{zhou2018interpreting}, by attaching OOD detectors to the intermediate layers, we could detect OODs based on features of different complexities. Figure \ref{fig:number_of_detected_ood} shows the number of OODs identified by different OOD detectors. For the LSUN and Tiny ImageNet OOD datasets which are of higher complexity, most of them are identified by the last two OOD detectors, while for the other three OOD datasets that have relatively lower complexity, they are mainly detected by the first seven detectors.

\begin{figure*}[!h]
    \vspace{-3mm}
    \centering
    \includegraphics[width=0.9\linewidth]{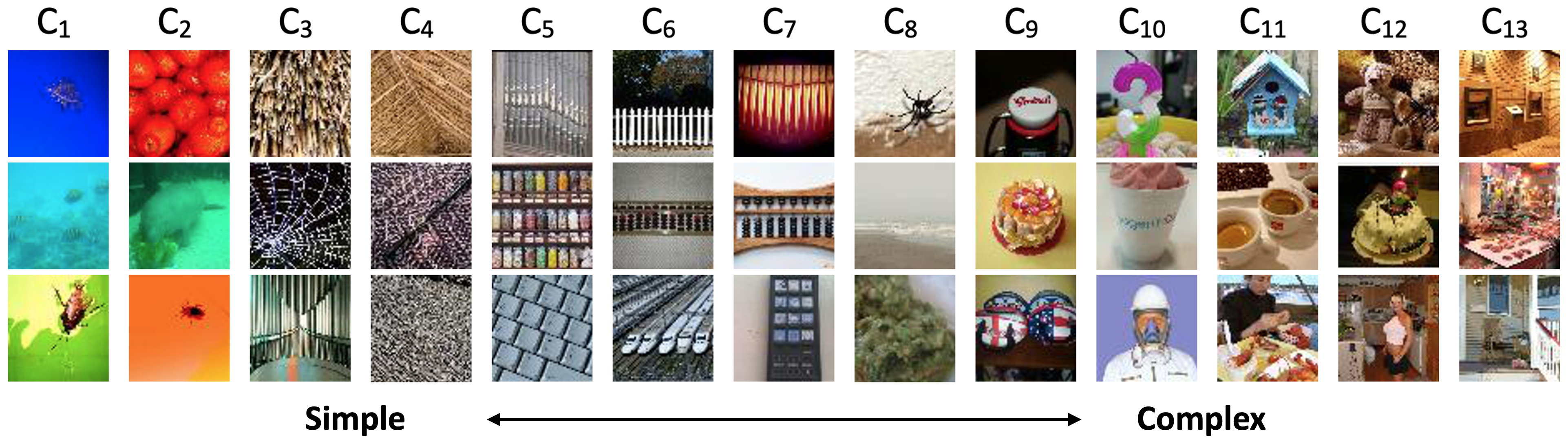}
    \vspace{-3mm}
    \caption{Correctly identified Tiny ImageNet OODs by OOD detectors at different layers, using VGG backbone and CIFAR10 as InD dataset.}
    \label{fig:simple_to_complex_tiny}
    \vspace{-4mm}
\end{figure*}

\begin{figure*}[!h]
    \vspace{-3mm}
    \centering
    \includegraphics[width=0.9\linewidth]{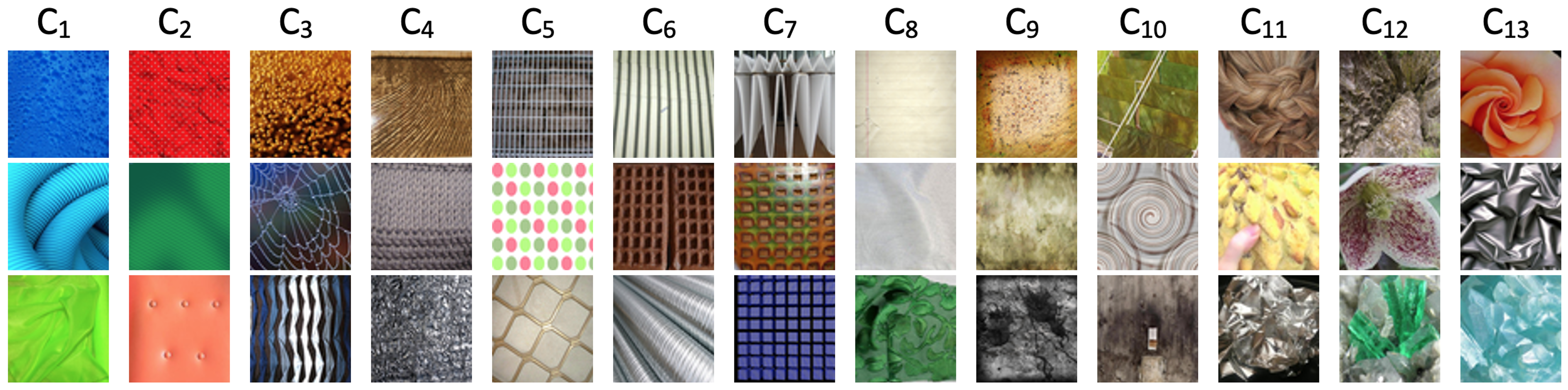}
    \vspace{-3mm}
    \caption{Correctly identified DTD OODs by OOD detectors at different layers, using VGG backbone and CIFAR10 as InD dataset.}
    \label{fig:simple_to_complex_dtd}
    \vspace{-5mm}
\end{figure*}

In Figure \ref{fig:simple_to_complex_tiny} we show the correctly identified Tiny ImageNet samples by different layer's OOD detectors using the VGG backbone and CIFAR10 as InD dataset. It could be seen that the OOD detectors at the initial layers are more sensitive to the image colors and textures which relate to the fine-scale details of the input images, while the OOD detectors at the ending layers tend to detect OODs based on objects or scenes. As the layer goes deeper, more and more complex OODs can be detected. Similar pattern could also be found on the DTD dataset, as shown in Figure \ref{fig:simple_to_complex_dtd}.

\subsection{Framework Confusion Analysis} \label{section:confusion}

\begin{wrapfigure}{r}{0.38\linewidth}
    \vspace{-18mm}
    \includegraphics[width=\linewidth]{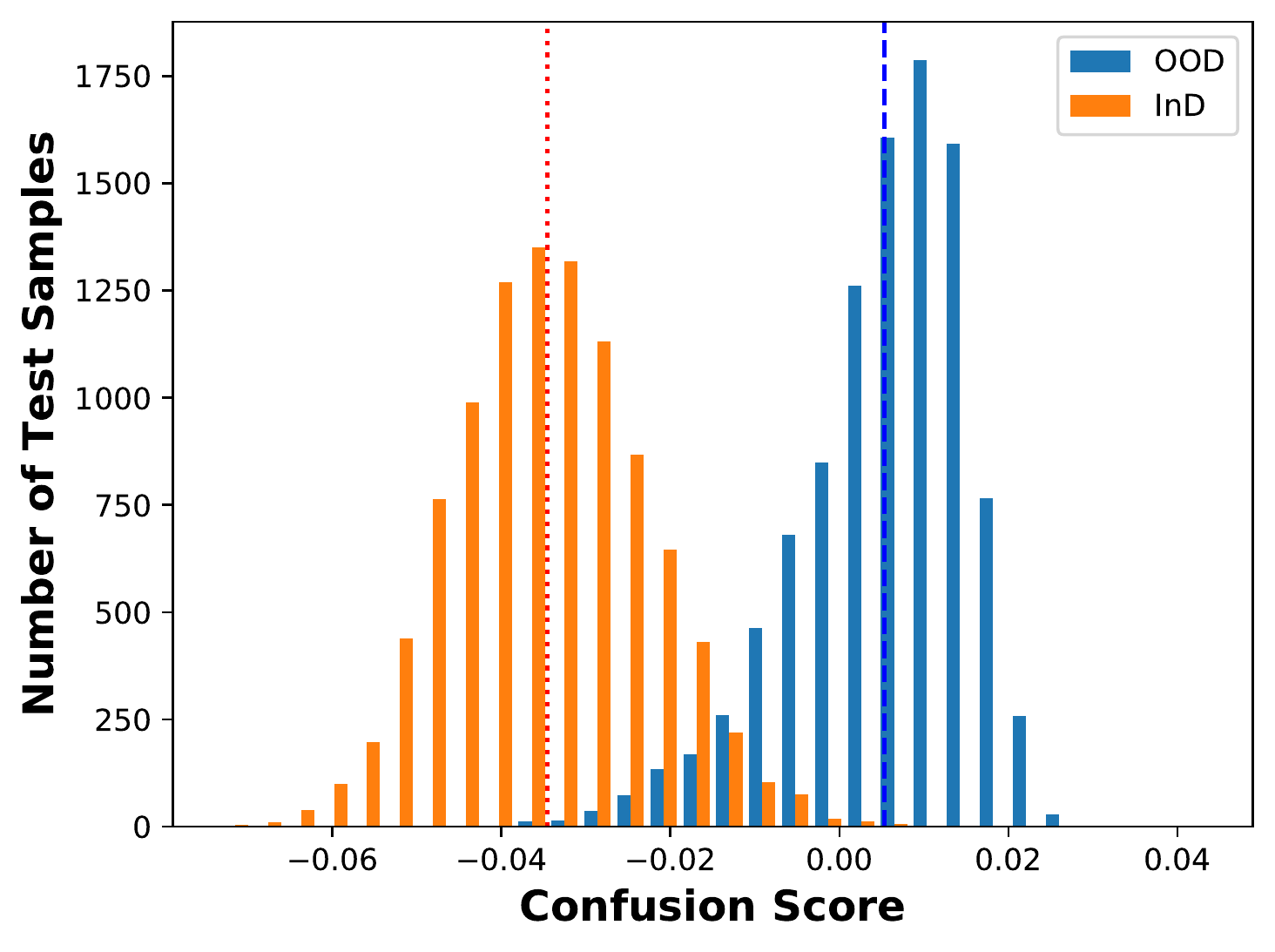}
    \vspace{-6mm}
    \caption{Confusion score of SVHN vs. CIFAR10 on VGG-16.}
    \label{fig:confusion_score}
    \vspace{-5mm}
\end{wrapfigure}

The disagreement between the OOD detectors indicates that their predictions are inconsistent and \textit{confused}. Here we define a confusion score $D(\mathbf{x})=\sum_1^L{C_{\ell}(\mathbf{x}^{(\ell)})}$ to measure the prediction divergence between the OOD detectors. For a good OOD detector, this confusion score should be negative for most of the InD test samples and positive for predicted OODs, the confusion occurs when the confusion score is close to 0.

We expect this confusion metric to be a reliable indicator in cases where the framework is unable to make a confident prediction and may have misclassified a test sample. Such an indicator has significant importance in handling errors due to the possible severe impact of false positives in real-world applications. We performed a confusion analysis on VGG backbone, using CIFAR10 as InD and SVHN as OOD, the confusion scores are shown in Figure \ref{fig:confusion_score}. While the InD samples tend to have small negative values (with an average of -0.16), the OOD samples are more concentrated on the positive side (with an average of 0.02). More importantly, the majority of the InD samples (99.78\%) have negative confusion scores and this makes the confusion analysis highly reliable and less prone to false positives. The confusion happens when the confusion score is close to zero, according to applications, a threshold could be determined based on the tolerance for misclassification.

\begin{wrapfigure}{r}{0.5\linewidth}
    \vspace{-5mm}
    \includegraphics[width=\linewidth]{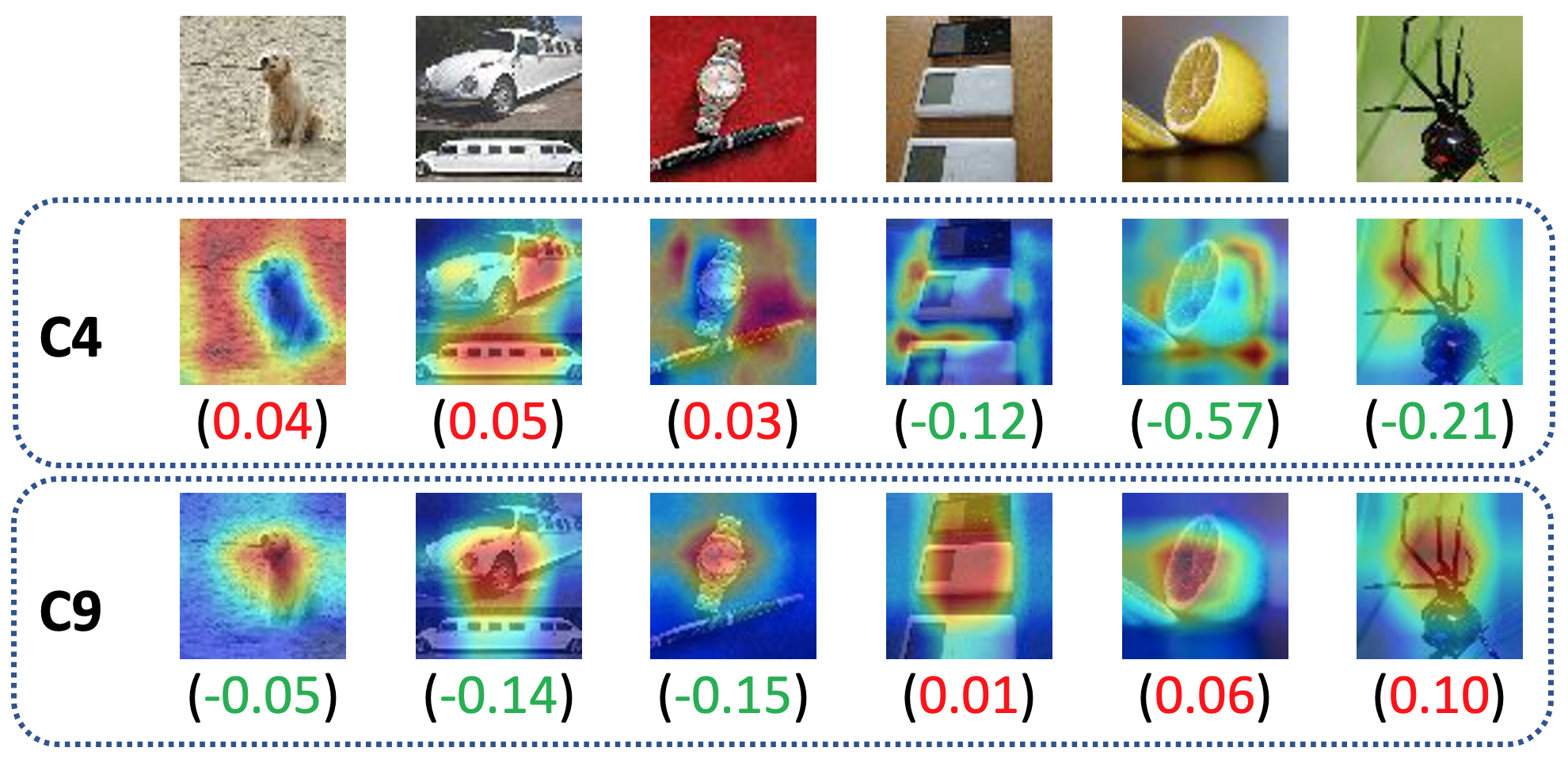}
    \vspace{-5mm}
    \caption{Prediction visualization of Tiny ImageNet samples, on VGG-16 and CIFAR10 InD.}
    \label{fig:grad_cam}
    \vspace{-5mm}
\end{wrapfigure}

Towards this error mitigation problem, we carry on the confusion analysis by designing a visualization tool for image OOD detection. Specifically, we adopt the Grad-CAM\cite{selvaraju2017grad} to show the root causes of the OOD predictions in the input space. The analysis is continued on the VGG backbone and CIFAR10 InD setting. As for the OOD dataset, we use the Tiny ImageNet since it has the most related class definition as CIFAR10. Some examples are shown in Figure \ref{fig:grad_cam} to illustrate the disagreement between two OOD detectors: C4 and C9, the numbers below the heatmaps are their corresponding OOD scores, with red color representing an OOD prediction and green color representing an InD prediction. We could see that OOD detectors at the early layers are more sensitive to textures and colors, while OOD detectors at the ending layers are more focused on objects and scenes.

\subsection{Advantages of Using Intermediate OOD Detectors}

\begin{wrapfigure}{r}{0.35\linewidth}
    \centering
        \centering
        \includegraphics[width=\linewidth]{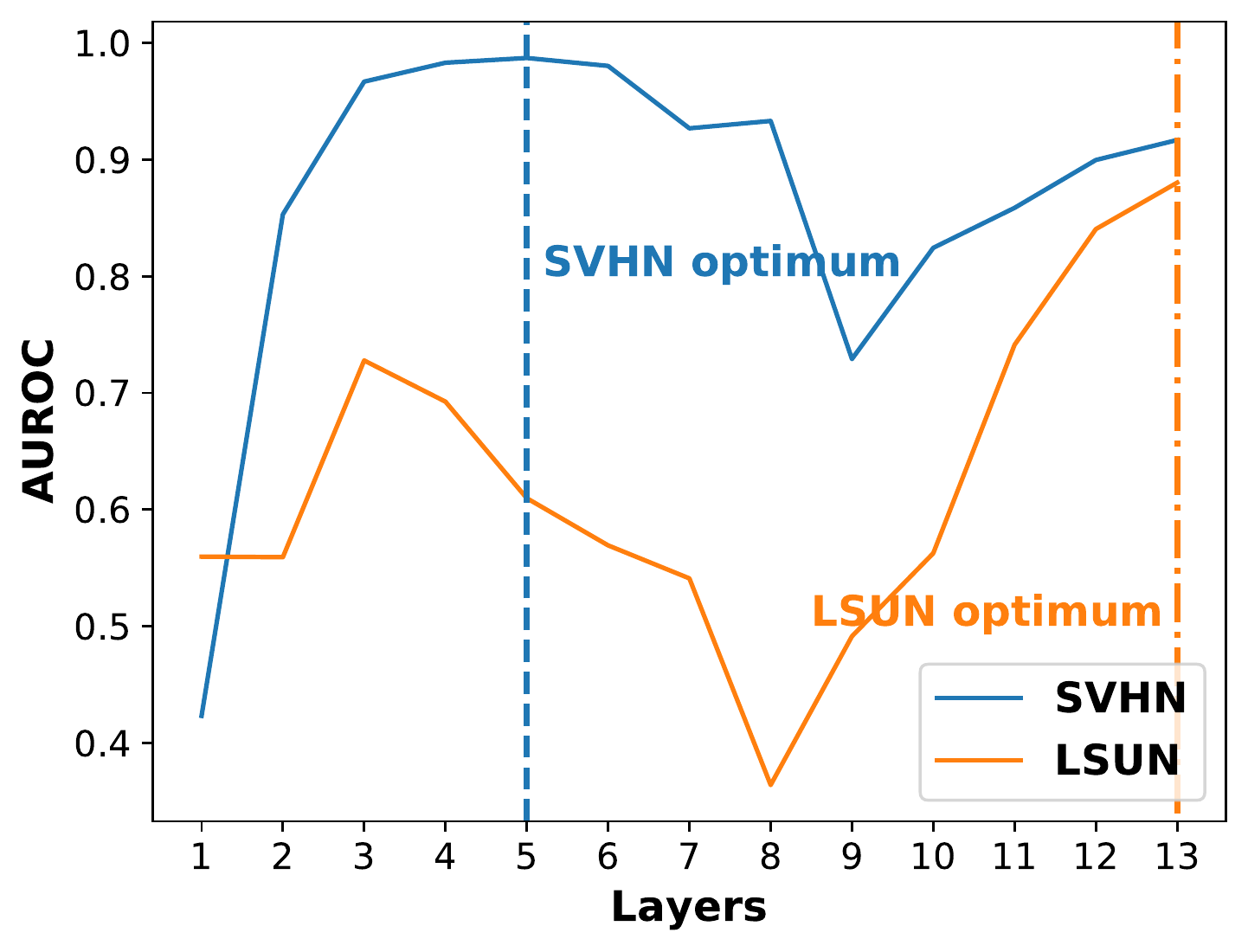}
    \vspace{-5mm}
    \caption{The optimal discernment layers of SVHN and LSUN on VGG-16.}
    \label{fig:optimal_layer}
    \vspace{-5mm}
\end{wrapfigure}

An optimal discernment layer\cite{abdelzad2019detecting} (or best layer) could be found for a particular OOD dataset, but it may not be the optimal choice for OOD datasets of different complexity. In Figure \ref{fig:optimal_layer} we show the AUROC of SVHN and LSUN at each layer of VGG-16 (using CIFAR10 as InD). The best layer for SVHN is layer 5, while the best layer for LSUN is the last layer. Such best layer could be estimated using a separate OOD dataset, however, as we could see from Table \ref{table:performance_evaluation}, OODL that estimates the best layer using the iSUN dataset could have its performance degrade significantly when OODs of different complexity are encountered. Therefore, instead of choosing the best layers for different OODs, \sysname{} propagates the most confident OOD prediction across all layers, and could effectively construct a good OOD confidence measurement for unseen OODs. For all five OOD datasets considered in this paper, \sysname{} can achieve competitive or even better accuracy compare to their corresponding best layers.

\subsection{Ablation Studies}

\begin{table*}[!th]
\vspace{-5mm}
\normalsize
\centering
\caption{Performance average on all the OOD datasets. Evaluation metrics with "$\uparrow$" indicate the bigger the better and "$\downarrow$" indicate the smaller the better. Best performance is labeled in \textbf{bold}.}
\setlength\tabcolsep{3.5pt}
\resizebox{\textwidth}{!}{
\begin{tabular}{cccccccccccccccc}
\hline
InD / Model                     & Metric          & $C_1$    & $C_2$    & $C_3$    & $C_4$    & $C_5$    & $C_6$    & $C_7$    & $C_8$    & $C_9$    & $C_{10}$   & $C_{11}$   & $C_{12}$   & $C_{13}$   & \sysname{}           \\ \hline
\multirow{4}{*}{\begin{tabular}[c]{@{}c@{}}CIFAR10\\      VGG-16\end{tabular}}  
                          & AUROC $\uparrow$           & 60.20 & 78.14 & 89.55 & 89.00 & 83.92 & 81.13 & 77.99 & 70.45 & 65.96 & 71.58 & 83.57 & 89.20 & 91.62 & \textbf{93.73} \\
                          & AUPR $\uparrow$        & 89.18 & 94.60 & 97.55 & 97.51 & 96.36 & 95.75 & 94.87 & 92.80 & 87.07 & 89.27 & 94.46 & 97.08 & 97.79 & \textbf{98.48} \\ 
                          & FPR at 95\% TPR $\downarrow$ & 95.47 & 84.39 & 61.76 & 64.55 & 77.97 & 85.61 & 89.12 & 95.19 & 88.47 & 82.01 & 56.04 & 63.22 & 37.16 & \textbf{28.25} \\\hline
\multirow{4}{*}{\begin{tabular}[c]{@{}c@{}}CIFAR100\\      VGG-16\end{tabular}} 
                          & AUROC$\uparrow$           & 51.87 & 70.09 & 83.71 & 82.61 & 79.72 & 77.08 & 76.17 & 69.55 & 72.43 & 70.38 & 62.66 & 37.80 & 73.46 & \textbf{85.34} \\
                          & AUPR$\uparrow$        & 85.56 & 91.89 & 95.93 & 95.85 & 95.19 & 94.51 & 93.84 & 91.90 & 91.11 & 90.27 & 86.70 & 76.10 & 90.42 & \textbf{95.93} \\ 
                          & FPR at 95\% TPR$\downarrow$ & 94.75 & 89.32 & 76.22 & 80.75 & 83.96 & 86.80 & 86.00 & 90.39 & 81.74 & 86.29 & 85.16 & 96.38 & 65.37 & \textbf{52.58} \\\hline
\end{tabular}}
\label{table:ablation_study}
\vspace{-6mm}
\end{table*}

Here we evaluate the effectiveness of the ``early exits". We compare the results of the proposed \sysname{} with the average performance using each OOD detector solely on five OOD datasets mixed (LUSN + Tiny ImageNet + SVHN + DTD + Pure Color). Results are shown in Table \ref{table:ablation_study}. Using VGG-16 as an example, for both CIFAR10 and CIFAR100 InD settings, \sysname{} can achieve consistently better performance than any single OOD detector.

\section{Conclusion}
    We proposed the \sysname{}, a layer-adaptive OOD detection framework for deep neural networks. By attaching multiple intermediate OOD detectors to the DNNs, \sysname{} can fully exploit the intrinsic characteristics of the intermediate latent space and reveal OODs with increasing complexity at deeper layers. Extensive experiments have been conducted to verify the effectiveness and interpretability of \sysname{}. On three DNNs with varying depth and architectures, our framework outperforms the state-of-the-art baselines without using any OOD training/validation data, being a reliable method for detecting unseen OODs.
%
%
%
\bibliographystyle{splncs04}
\bibliography{references.bib}
%

\end{document}